\documentclass[runningheads]{llncs}
\usepackage[T1]{fontenc}
\usepackage{graphicx}
\usepackage{amsmath}
\usepackage{algorithm}
\usepackage{algorithmic}
\usepackage{amsfonts}
\usepackage{listings}
\usepackage{xcolor}
\usepackage{hyperref} 
\definecolor{codegray}{gray}{0.9}
\definecolor{keywordcolor}{RGB}{255, 0, 0}
\definecolor{commentcolor}{RGB}{0, 128, 0}
\definecolor{stringcolor}{RGB}{0, 0, 255}

\begin{document}
\renewcommand{\thefootnote}{\fnsymbol{footnote}}  
\title{Sine Wave Normalization for Deep Learning-Based Tumor Segmentation in CT/PET Imaging}
%
%
\author{Jintao Ren\thanks{Corresponding} \inst{1,2}\orcidID{0000-0002-1558-7196} \and 
Muheng Li\inst{3}\orcidID{0000-0002-0619-5780} \and
Stine Sofia Korreman\inst{1,2}\orcidID{0000-0002-3523-382X} }
\authorrunning{J. Ren et al.}
%
\institute{Aarhus University, Department of Oncology, Nordre Ringgade 1,
8000 Aarhus, Denmark \and
Aarhus University Hospital, Danish Centre for Particle Therapy, Palle Juul-Jensens Blvd. 25, 8200 Aarhus, Denmark \and
Paul Scherrer Institute PSI, Center for Proton Therapy, Forschungsstrasse 111 5232 Villigen PSI, Switzerland
\\ \email{jintaoren@clin.au.dk},
\\ 
}
\maketitle              
\begin{abstract}
This report presents a normalization block for automated tumor segmentation in CT/PET scans, developed for the autoPET III Challenge. The key innovation is the introduction of the SineNormal, which applies periodic sine transformations to PET data to enhance lesion detection. By highlighting intensity variations and producing concentric ring patterns in PET highlighted regions, the model aims to improve segmentation accuracy, particularly for challenging multitracer PET datasets. The code for this project is available on \href{https://github.com/BBQtime/Sine-Wave-Normalization-for-Deep-Learning-Based-Tumor-Segmentation-in-CT-PET}{GitHub}.

\end{abstract}
\section{Introduction}
Positron Emission Tomography (PET) combined with Computed Tomography (PET/CT) is a vital tool for diagnosing, managing, and treating oncological diseases. PET/CT provides both anatomical and metabolic insights, allowing precise tumor localization and characterization. However, PET intensity values vary significantly due to factors like tracer type, patient metabolism, imaging protocols, and scanner sensitivity, complicating lesion detection and segmentation. Normalizing PET data is essential to reduce this variability and ensure consistent input for deep learning models. Accurate tumor segmentation remains a primary challenge, crucial for quantitative analysis and personalized therapy planning.

The autoPET III Challenge \cite{Ingrisch2024} addresses this issue by providing a multitracer, multicenter dataset, focusing on the development of robust deep learning models that can generalize across different tracers (e.g., PSMA and FDG). The challenge aims to advance automated lesion segmentation.

This report introduces a deep learning-based approach to address the AutoPET III challenge of automatic lesion segmentation in PET/CT images. Building upon the nnUNet ResEnc(M) plan \cite{isensee2021nnu,isensee2024nnu} and UMamba block \cite{ma2024u}, we propose a modification specifically tailored for PET data: the SineNormal. This component is designed to normalize PET data while capturing and emphasizing subtle metabolic variations, potentially enhancing the model’s capacity to detect and segment tumor lesions. While the approach appears promising, comprehensive validation is still necessary to confirm its effectiveness.

\section{Method}
\subsection{Sine Normal Transformation}
A key contribution of this work is the introduction of the \texttt{SineNormal}, a module specifically designed to enhance PET data processing in the context of tumor segmentation. PET images, which represent metabolic activity, often exhibit complex intensity patterns. The \texttt{SineNormal} applies periodic sine wave transformations to the PET input, aiming to capture subtle variations in intensity and spatial features that are critical for accurate lesion detection.

The primary function of the \texttt{SineNormal} is to amplify these metabolic variations through non-linear transformations. By leveraging sine functions with different frequency constants, the module highlights both small and large-scale intensity gradients, which may correspond to tumor boundaries and metabolic heterogeneities within the lesion.

Given an input tensor $\mathbf{x}_{\text{PET}} \in \mathbb{R}^{D \times H \times W}$, where $D$, $H$, and $W$ are the spatial dimensions of the PET image, the transformation applied by the \texttt{SineNormal} can be described as follows:
\begin{equation}
    \mathbf{y} = \sin(\mathbf{a} \cdot \mathbf{x}_{\text{PET}}),
\end{equation}
here, $\mathbf{a}$ is a vector of constants $\mathbf{a} = [a_1, a_2, \dots, a_c]$ where $c$ represents the number of channels after the input is repeated. The sine function is applied element-wise to the PET image data, where each constant $a_i$ corresponds to a different frequency component. This transformation introduces oscillations into the image, creating concentric ring patterns around high-intensity regions, such as tumor cores, thereby enhancing contrast and highlighting critical features \cite{ren2021pet}.

\subsection{UNet architecture}
The network architecture is based on the nnUNet ResEnc(M) plan, which employs a residual encoder \cite{he2016deep} and normal CNN decoder structure to extract and reconstruct features for segmentation tasks. The network consists of 6 stages, with progressively increasing feature channels to capture increasingly abstract representations of the input. The feature sizes for each stage are set to [32, 64, 128, 256, 320, 320]. Throughout the network, 3D CNNs are used to process volumetric data, with a kernel size of [3, 3, 3]. The decoder uses one convolution per stage to upsample the encoded features and reconstruct the segmented output. Deep supervision was enabled. The number of blocks per stage is configured as [1, 3, 4, 6, 6, 6]. To improve feature representation before each skip connection, a UMamba block was added after every CNN encoder stage except the first. This design avoids excessive computational burden in the first stage while leveraging UMamba's ability to capture long-range dependency at deeper stages\cite{liu2024swin,wang2024mamba}.

\subsection{SineNormalBlock}
The SineNormalBlock was applied to the input PET channel after 0-1 normalization, generating two sine-transformed channels with constants $a = 20$ and $a = 30$. These transformed channels were concatenated with the original PET and CT inputs, resulting in four channels (CT, PET, and two sine-normalized PET channels) being fed into the U-Net encoder.

The following code defines the \texttt{SineNormalBlock} in PyTorch:
\begin{lstlisting}[language=Python, caption={SineNormalBlock Code in PyTorch. The variable 'constant\_a' can be customized for different frequency patterns. In this case, the values 20 and 30 were chosen because, after 0-1 normalization, a factor of 20 generates sufficient concentric rings, while 30 introduces a slightly different pattern for additional variation.}]
class SineNormalBlock(nn.Module):
    def __init__(self, hidden_channels=2):
        super(SineNormalBlock, self).__init__()
        self.hidden_channels = hidden_channels
        # Constant values for the sine activation
        self.constant_a = torch.tensor([20.0, 30.0])

    def forward(self, x):
        # Repeat input along the channel dimension
        x = x.repeat(1, self.hidden_channels, 1, 1, 1)
        # Ensure constant_a is on the same device as input tensor x
        self.constant_a = self.constant_a.to(x.device)
        # Apply sine transformation
        x = torch.sin(self.constant_a.view(1, -1, 1, 1, 1) * x)

        return x

\end{lstlisting}

\subsection{Training configurations}
The entire training dataset (n=1611) was used without splitting. The network was trained with a batch size of 8 and a patch size of $112\times160\times128$ voxels. CT images were first clipped at the 0.05–99.5 percentile and then normalized using z-score scaling, while PET images were normalized to a 0-1 range. The training was conducted using the SGD optimizer with a PolyLR scheduler (exponent = 0.9) starting at a learning rate of 0.01. The loss function was a combination of cross-entropy and Dice loss. Gradient accumulation was applied over 8 steps. Training was conducted for up to 1100 epochs, with the best-performing model checkpoint obtained around epoch 1050.

\section{Post Processing}
AutoPET III imposes a 5-minute limit per patient. To maximize computational efficiency, we implemented both a dynamic sliding window approach and dynamic test-time augmentation (TTA). For the sliding window approach, we initialized the step size at 0.5 for all axes. In the coronal and sagittal planes, steps were limited to a maximum of 4, starting from the image center. Step sizes in these planes were dynamically adjusted based on coverage, increasing by 0.1 per iteration until coverage exceeded 80\%. 

Additionally, TTA was adjusted based on the number of sliding window steps in the axial direction. When the axial plane had 8 steps, only half of the mirroring axes were applied. For cases with more than 8 steps, mirroring was reduced to one-quarter of the mirroring axes. For cases with fewer than 8 steps, all axes were used for mirroring in the TTA.

\section{Discussion and Future Work}
The goal of this report was to present our approach to tumor segmentation using sine wave normalization in CT/PET imaging. The effectiveness of the approach was not tested due to time constraints. The method should be further thoroughly evaluated using proper data splitting and validation. Future work will involve rigorous evaluation to assess the model's performance and potential for clinical application.

\bibliographystyle{splncs04}  
\bibliography{references}  
\end{document}